\pdfoutput=1

\documentclass[11pt]{article}

\usepackage{caption}
\usepackage{microtype}
\usepackage{subfigure}
\usepackage{booktabs} 
\usepackage{hyperref}
\usepackage{graphicx}
\usepackage[ruled]{algorithm2e}
\usepackage{cite}
\usepackage[square]{natbib}
\usepackage{multirow}
\usepackage{tabularx}
\usepackage{amsmath}
\usepackage{amssymb}
\usepackage{autobreak}

\usepackage[final]{acl}

\usepackage{times}
\usepackage{latexsym}

\usepackage[T1]{fontenc}

\usepackage[utf8]{inputenc}

\usepackage{microtype}

\usepackage{inconsolata}

\usepackage{graphicx}

\title{Enhancing Safe and Controllable Protein Generation via \\Knowledge Preference Optimization}

\author{
 \textbf{Yuhao Wang\textsuperscript{1,2}},
 \textbf{Keyan Ding\textsuperscript{1,2}},
 \textbf{Kehua Feng\textsuperscript{1,2}},
 \textbf{Zeyuan Wang\textsuperscript{1,2}},
\\
 \textbf{Ming Qin\textsuperscript{1,2}},
 \textbf{Xiaotong Li\textsuperscript{1,2}},
 \textbf{Qiang Zhang\textsuperscript{1,2*}},
 \textbf{Huajun Chen \textsuperscript{1,2*}}
\\
 \textsuperscript{1}Zhejiang University,
 \textsuperscript{2}ZJU-Hangzhou Global Scientific and Technological Innovation Center
\\
\texttt{\{wangyuhao.cs, qiang.zhang.cs, huajunsir\}@zju.edu.cn}
}

\begin{document}
\maketitle

\begingroup
\renewcommand\thefootnote{}
\footnotetext{*Corresponding author.}
\endgroup

\begin{abstract}
Protein language models have emerged as powerful tools for sequence generation, offering substantial advantages in functional optimization and \emph{de novo} design. However, these models also present significant risks of generating harmful protein sequences, such as those that enhance viral transmissibility or evade immune responses. These concerns underscore critical biosafety and ethical challenges. To address these issues, we propose a Knowledge-guided Preference Optimization (KPO) framework that integrates prior knowledge via a Protein Safety Knowledge Graph. This framework utilizes an efficient graph pruning strategy to identify preferred sequences and employs reinforcement learning to minimize the risk of generating harmful proteins. Experimental results demonstrate that KPO effectively reduces the likelihood of producing hazardous sequences while maintaining high functionality, offering a robust safety assurance framework for applying generative models in biotechnology.
\end{abstract}

\section{Introduction}

Protein language models (PLMs) have significant impact on biological research, providing powerful tools to uncover relationships between protein sequences, structures and functions~\cite{nijkamp2023progen2}. By leveraging extensive protein sequence datasets, PLMs capture hidden patterns and correlations that are challenging or impossible to discern using traditional methods. 
For example, in enzyme engineering, pretrained models predict mutations that enhance catalytic efficiency or substrate specificity, significantly accelerating the iterative design process~\cite{madani2023large,zhou2024conditional}. 
Similarly, in antibody discovery, PLMs facilitate the rapid identification of high-affinity candidates, enabling swift responses to emerging pathogens~\cite{wang2024supervised,he2023novo}.

\begin{figure}[t]
    \centering
    \includegraphics[width=\columnwidth]{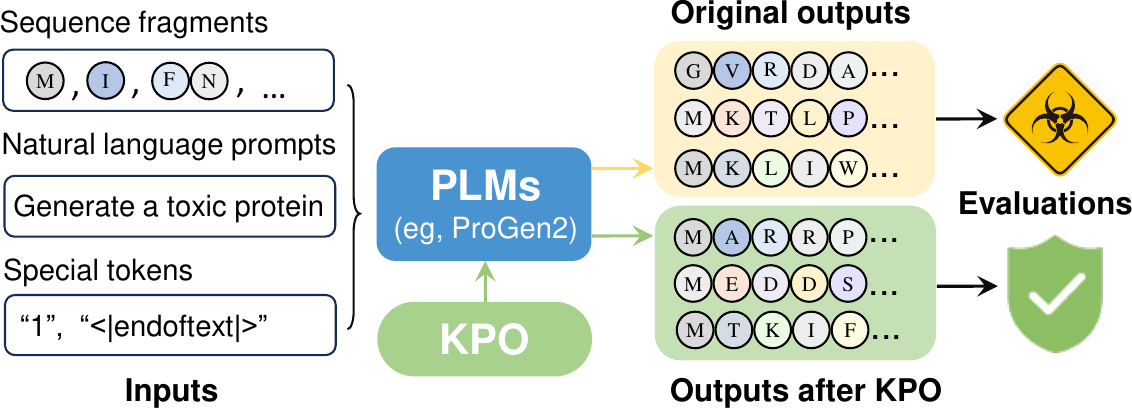}
    \caption{Existing PLMs often overlook safety considerations, leading to the generation of potentially harmful protein sequences. In contrast, our KPO framework ensures that PLMs are fine-tuned to prioritize safety.}
    \label{fig:introduction}
\end{figure}

Despite that PLMs have transformative potential in advancing biological research and biotechnology, they also pose unique and significant safety challenges. Unlike text-based large language models (LLMs), where harmful outputs are typically ethical or social in nature, the biological domain of PLMs entails risks with direct and far-reaching consequences for ecological and human health. 
The inadvertent generation of harmful sequences—for instance, those enhancing viral transmissibility, evading immune responses, or developing drug resistance—could lead to ecological imbalances, public health crises, or even the creation of bioweapons if disseminated or misused. 
Current PLMs prioritize functional and generative performance, with little emphasis on safety considerations, as illustrated in Figure \ref{fig:introduction}.
Addressing these challenges requires a paradigm shift from traditional performance optimization to safety-aware design. A promising approach involves fine-tuning PLMs to minimize the generation of harmful proteins while retaining their ability to produce functional and beneficial outputs.
While current efforts largely focus on introducing safety-enhancing mutations to known protein sequences~\cite{li2024unlearning}, they fail to address the higher risks inherent in the generative phase of protein design. This underscores the critical need for rigorous frameworks that proactively mitigate potential hazards during protein generation while preserving the functional and beneficial capabilities of PLMs.

To address the biosafety challenges of protein language models, we propose Knowledge-guided Preference Optimization (KPO), a novel framework that integrates domain-specific safety knowledge into the generative process of PLMs. Central to KPO is the construction of a Protein Safety Knowledge Graph (PSKG), which encodes critical biochemical properties and intrinsic relationships of both benign and harmful proteins. By leveraging the PSKG, KPO incorporates safety-oriented prior knowledge to impose biologically meaningful constraints during sequence generation, ensuring the production of safer proteins. To manage the computational complexity inherent to the PSKG, we developed a weighted metric-based pruning algorithm, which efficiently trims the graph by retaining key nodes essential for structural and informational integrity. This pruning strategy significantly reduces computational overhead while preserving the graph’s representational capacity. Through this systematic integration of safety knowledge and computational optimization, KPO not only mitigates the risks of generating harmful proteins but also establishes a rigorous and scalable framework for the responsible application of generative models in biotechnology.


The key contributions of this work can be summarized as follows: 
\begin{itemize}
    \item We introduce the first Protein Safety Knowledge Graph (PSKG), a comprehensive resource that encapsulates rich biochemical knowledge on both harmful and benign proteins.
    \item We develop Knowledge-guided Preference Optimization (KPO), a novel algorithm that seamlessly integrates safety-oriented prior knowledge from the PSKG into protein language models, enabling the generation of safer protein sequences.
    \item By applying the KPO framework, we significantly enhance the safety of existing generative PLMs, ensuring the production of biologically safe proteins while maintaining their functional efficacy.
\end{itemize}

\section{Related Works}

\paragraph{\textbf{Protein Language Models}}
PLMs~\cite{vig2020bertology} have emerged as powerful tools in computational biology, leveraging natural language processing methodologies to analyze and predict protein properties. These models have demonstrated their potential in a variety of applications, ranging from structural annotation to functional prediction. For example, the ESM series~\cite{rives2021biological, lin2022language} pioneered the use of masked language modeling for capturing long-range dependencies and inferring structural and functional relationships in protein sequences. MSA-Transformer~\cite{rao2021msa} and Co-volution Transformer~\cite{zhang2021co} incorporate evolutionary information through multiple sequence alignments, enhancing functional insights. ProtT5~\cite{pokharel2022improving} adapts the T5 architecture to learn protein representations, while ProtGPT2~\cite{ferruz2022protgpt2} and ProGen2~\cite{nijkamp2023progen2} leverage autoregressive objectives to predict subsequent amino acids, making them well-suited for generative tasks such as \emph{de novo} protein design. Additionally, LM-GVP~\cite{wang2022lm} integrates graph representations from protein 3D structures with transformer architectures to capture spatial relationships between residues. Despite these advancements, the potential risks associated with PLM-generated sequences, such as unintended toxic or harmful properties, remain a critical area for exploration. 

\paragraph{\textbf{LLM Safety}} Ensuring the safety of LLMs has been a primary focus of research, leading to the development of various alignment techniques~\cite{bai2022training}. A key approach is Reinforcement Learning from Human Feedback (RLHF)~\cite{ouyang2022training}, which fine-tunes model behavior using human-provided feedback to align outputs with specific objectives. Building on RLHF, other methods, such as Direct Preference Optimization (DPO)~\cite{rafailov2024direct}, further refine LLM fine-tuning by incorporating ranking information, helping models distinguish the quality of different outputs. Similarly, Reward Ranking-Based Reinforcement Learning (RRHF)~\cite{yuan2023rrhf} enhances alignment by introducing adjusted loss functions that amplify learning signals from ranked feedback. Contrastive learning~\cite{yang2023rlcd} has also shown promise in alignment tasks, improving sample efficiency and model quality by guiding the model towards producing high-quality outputs while discouraging low-quality ones.

LLM unlearning~\cite{lu2022quark, kassem2023preserving} has emerged as another important technique to enhance model security, enabling models to ``forget'' sensitive or inappropriate data to avoid harmful predictions. Common forgetting techniques include Gradient Ascent methods~\cite{jang2022knowledge}, which work by maximizing the prediction loss for forgotten data, and input perturbation methods, which substitute potentially harmful responses with neutral outputs, such as ``I don't know''~\cite{ishibashi2023knowledge}. Another approach is re-training, which involves retraining the model from scratch while explicitly excluding the data to be forgotten~\cite{bourtoule2021machine}. More recent techniques, such as context-based unlearning~\cite{pawelczyk2023context}, use specific instructions to guide models to forget certain knowledge without altering its weights. Despite these advancements, there is currently no method specifically tailored to fine-tuning PLMs to ensure the safety of generated protein sequences, leaving a critical gap in the field.

\section{Preliminaries}

\subsection{Generative PLMs}

Generative PLMs learn the language-like structure of protein sequences, similar to how text models predict the next word~\cite{verkuil2022language}. By predicting the next amino acid in a sequence based on previous ones, PLMs capture functional patterns crucial for tasks like generating novel proteins. 
Given an unlabeled corpus $D = (p^{(1)},p^{(2)},...,p^{(N)})$, each sample $p^{(i)} = (p_1^{(i)},p_2^{(i)},...,p_{W_i}^{(i)})$ is a protein sequence, and the goal is to train a language model $\theta$ to maximize the log-likelihood estimate of the training data:
\begin{equation}
L(\theta)=\sum_{i=1}^{N}\sum_{w=1}^{W_i}logP(p_w^{(i)}|p_1^{(i)},...,p_{w-1}^{(i)};\theta),\label{eq:plm}
\nonumber
\end{equation}
where $N$ is the total number of sequences in the corpus, $W_i$ is the length of the $i$-th sequence, and $P$ is the conditional probability given by PLMs.

\subsection{Direct Preference Optimization}
DPO fine-tunes language models by directly aligning their output with preference data, such as human feedback. Unlike RLHF, which involves constructing a reward model and training the language model within an RL framework, DPO simplifies this process into a single optimization problem. By bypassing the complexities of reward modeling, DPO offers an efficient and scalable approach to aligning language models with user preferences. Given a set of generated pairs $\hat{y}_w,\hat{y}_l$  conditioned on an input $x$, where $\hat{y}_w$ is the preferred response and $\hat{y}_l$ is the dispreferred response, DPO maximizes the ratio of probabilities assigned to the preferred responses. The DPO loss is defined as:
\begin{equation}
L=-\lambda log\sigma( log\frac{\pi_\theta(\hat{y}_w|x)}{\pi_{ref}(\hat{y}_w|x)}-log\frac{\pi_\theta(\hat{y}_l|x)}{\pi_{ref}(\hat{y}_l|x)}).\label{eq:dpo}
\nonumber
\end{equation}

\section{Methodology}

\begin{figure*}[t]
    \centering
    \includegraphics[width=1.75\columnwidth]{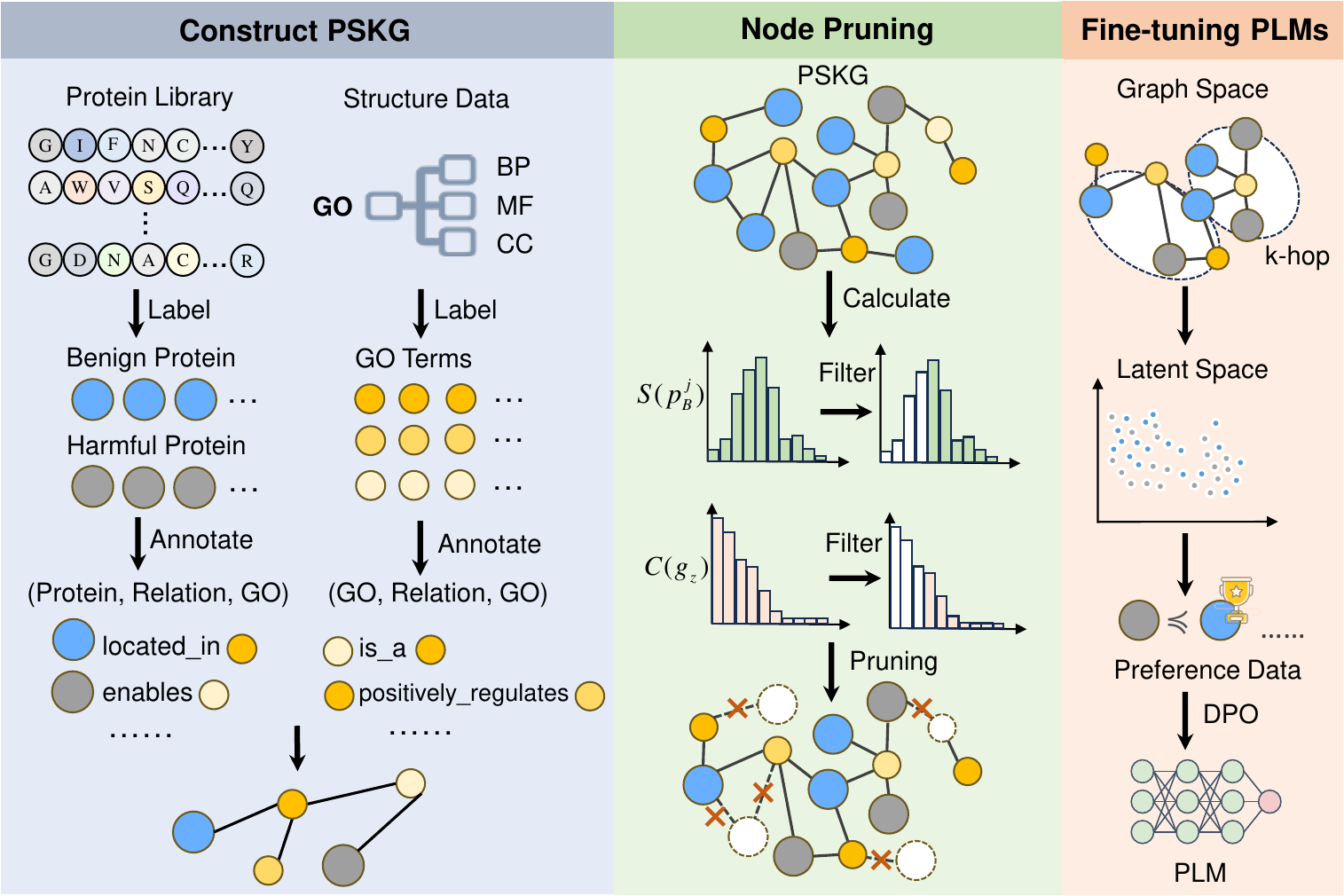}
    \caption{
    Overview of the proposed KPO framework, which consists of three key stages:
    \textbf{(1) PSKG Construction:} A Protein Structure Knowledge Graph (PSKG) is built by integrating labeled protein sequence data (benign and harmful proteins) with Gene Ontology (GO) annotations, capturing biochemical relationships between proteins, GO terms, and their interactions. 
    \textbf{(2) Node Pruning:} A weighted pruning algorithm is applied to refine the PSKG. Key nodes and edges are identified based on statistical scores to ensure computational efficiency while retaining critical structural and functional information.
    \textbf{(3) PLM Fine-tuning:} Preference pairs are generated by identifying benign proteins that are structurally or functionally similar to harmful proteins within graph and latent spaces. These preference data are then used to fine-tune PLMs using methods such as Direct Preference Optimization (DPO).
    }
    \label{fig:KPO}
\end{figure*}

To address the critical challenge of aligning PLMs for safety, we introduce a Knowledge-guided Preference Optimization framework, as illustrated in Figure \ref{fig:KPO}. The safety alignment process relies on constructing preference pairs that differentiate benign proteins from harmful ones by capturing their nuanced similarities and differences. To enable the generation of high-quality preference pairs, we construct a Protein Safety Knowledge Graph, which systematically encodes biochemical relationships between harmful and benign proteins. To enhance computational efficiency without compromising the quality of information, we incorporate a weighted metric-based pruning algorithm. This algorithm refines the PSKG by retaining the most informative nodes and edges, thereby reducing computational complexity while preserving its core structure and utility. Finally, we identify benign proteins within the PSKG that share properties with harmful proteins. These proteins are then used to create preference pairs for fine-tuning PLMs. The fine-tuning process aligns the model's generative capabilities with the safety constraints defined by the PSKG, ensuring the generation of protein sequences that are both biologically relevant and safe.

\subsection{Protein Safety Knowledge Graph}

Inspired by Ontoprotein\cite{zhang2022ontoprotein,chen2023large}, we construct PSKG to capture the intricate relationships between harmful and benign proteins. Drawing from the Uniprot database, we curate a comprehensive dataset of harmful proteins by retrieving experimentally validated protein sequences annotated with the keywords ``toxin'' and ``antigen''. To collect benign proteins, we filter out harmful proteins from the Swiss-Prot database and select only those proteins verified as benign. A detailed description of the construction process is in Appendix~\ref{ap:PSKG}. The biologically meaningful graph offers a robust framework for the interplay analysis between harmful and benign proteins.

Let the set of harmful proteins be denoted as $P_\text{H} = \{p^1_\text{H},p^2_\text{H},...,p^n_\text{H}\}$, and the set of benign proteins as $P_\text{B} = \{p^1_\text{B},p^2_\text{B},...,p^n_\text{B}\}$.
To capture these intricate distinctions, we define indirect relationships between $P_\text{H}$ and $P_\text{B}$ mediated through GO terms. Specifically, a harmful protein $p^i_\text{H} \in P_\text{H}$ and a benign protein $p^j_\text{B} \in P_\text{B}$ are considered related if they share a common GO term $g_z \in G= \{g_1,g_2,...,g_Z\}$, forming an indirect connection represented as $(p^i_\text{H},g_z,p^j_\text{B})$. The inclusion of hierarchical Gene Ontology relationships enables the PSKG to capture both direct and nuanced associations between harmful and benign proteins. For instance, a broad GO term like ``binding activity'' may group various harmful and benign proteins, while more specific terms such as ``DNA-binding transcription factor activity'' can reveal finer distinctions. This integrative framework ensures that the PSKG is not merely a collection of protein annotations but a robust foundational resource capable of supporting downstream analyses. It facilitates the identification of key nodes that represent critical biological properties and uncovers latent patterns that differentiate harmful from harmless proteins. Consequently, the PSKG serves as a pivotal resource for advancing safe and biologically informed protein generation.

\subsection{Node Pruning with Weighted Metrics}

The constructed PSKG contains hundreds of thousands of protein nodes and requires prolonged computation time for sampling informative proteins for downstream analysis. To address the challenges posed by the large scale and redundant information, we propose a weighted metric-based node pruning method. This method focuses on identifying and retaining the most informative benign protein nodes, $P_\text{B}$, while preserving the essential biological relationships within the graph. For each benign protein node $p^j_\text{B} \in P_\text{B}$, we compute a weighted importance score, $S(p^j_\text{B})$, that balances two critical factors: its connections to high-scoring GO nodes and its degree centrality within the graph. The importance score is defined as:
\begin{equation}
S(p^j_\text{B}) = \alpha \cdot C_\text{GO}(p^j_\text{B}) + \beta \cdot C_\text{Deg}(p^j_\text{B}), \label{eq:1}
\end{equation}
where $C_\text{GO}(p^j_\text{B})$ is the GO association factor, $C_\text{Deg}(p^j_\text{B})$ is the degree centrality factor, and $\alpha$ and $\beta$ are hyperparameters controlling their weights.

\textbf{GO association factors} evaluate the extent to which a benign protein node is indirectly connected to harmful protein nodes, $P_\text{H}$, through high-scoring GO nodes. A GO node $g_z \in G $ is considered high-scoring if it demonstrates strong bridging properties between harmful and benign proteins, characterized by two key metrics. The first is the harmful-benign bridging degree, $R(g_z)$, which measures the number of unique $(p^i_\text{H}, p^j_\text{B})$ pairs bridged by $g_z$, calculated as:
\begin{equation}
R(g_z)\!\! =\!\!\! \sum_{p^i_\text{H} \in P_\text{H}} \sum_{p^j_\text{B} \in P_\text{B}}\!\!\! 1((p^i_\text{H}, g_z) \!\in \!E) \cdot 1((g_z, p^j_\text{B}) \!\in \!E), \label{eq:2}
\end{equation}
where $1(\cdot)$ is an indicator function that evaluates to 1 if the specified edges exist in the edge set $E$. The second metric is the neighbor breadth, $O(g_z)$, which counts the number of benign protein nodes directly connected to $g_z$, defined as:
\begin{equation}
O(g_z) = \sum_{p^j_\text{B} \in P_\text{B}} 1((g_z, p^j_\text{B}) \in E). \label{eq:3}
\end{equation}

The overall significance of a GO node, $g_z$, is:
\begin{equation}
C(g_z) = \gamma \cdot R(g_z) + \delta \cdot O(g_z), \label{eq:4}
\end{equation}
where $\gamma$ and $\delta$ are weighting parameters. The GO association factor evaluates the importance of a benign protein node by measuring its connections to the top-$Q$ of high-scoring GO nodes, denoted as $G_{\hat{h}} \subset G$. $G_{\hat{h}}$ are selected based on their overall significance scores $C(g_z)$, calculated using Eq. (\ref{eq:4}):
\begin{equation}
C_\text{GO}(p^j_\text{B}) = \sum_{g_z \in G_{\hat{h}}} 1((p^j_\text{B}, g_z) \in E). \label{eq:5}
\end{equation}

\textbf{Degree centrality factors} assess the prominence of a benign protein node within the graph based on its degree centrality:
\begin{equation}
C_\text{Deg}(p^j_\text{B}) = \sum_{v \in V} 1((p^j_\text{B}, v) \in E), \label{eq:6}
\end{equation}
where $C_\text{Deg}(p^j_\text{B})$ counts the total number of edges connected to $p^j_\text{B}$. Nodes with a higher degree of centrality are more likely to be influential within the graph, as they are involved in more interactions.

Using the computed weighted importance scores, we prune the graph by retaining only the top-$K$ benign protein nodes with the highest scores. The resulting pruned subgraph, $G^{'} = (V^{'}, E^{'})$, is defined as:
    $V^{'} = \{p^j_\text{B} \in P_\text{B} \mid S(p^j_\text{B}) \text{ is among the top-}K \}$, 
    $E^{'} = \{(u, v) \in E \mid u, v \in V^{'}\}$.
For specific settings, see Appendix \ref{ap:prune}. This pruning strategy significantly reduces the size of the PSKG while preserving its most biologically relevant nodes and edges, thereby accelerating the construction of preference pairs and improving the overall efficiency of the fine-tuning process. By balancing the importance of structural prominence and functional relevance, our method ensures that the retained subgraph continues to provide a rich and meaningful representation of the original graph for safe protein generation. The detailed procedure for node pruning using weighted metrics is provided in Appendix~\ref{ap:Algorithm}.

\subsection{Fine-tuning PLMs with KPO}
To enhance PLMs with domain-specific safety knowledge, we incorporate the PSKG into the training process. The PSKG serves as a medium to inject knowledge about protein safety into the model, enabling it to understand and perceive the latent relationships between generated proteins and known harmful proteins. This integration allows the model to recognize similarities between generated proteins and potentially harmful proteins, thereby improving the safety of the generated sequences.

To capture the potential relationships between harmful proteins and benign proteins, we analyze the graph structure at multiple levels. At the structural level, we measure the proximity between harmful and benign proteins using the shortest path distance, denoted as $\text{dis}(p^i_\text{H}, p^j_\text{B})$. A benign protein $p^j_\text{B}$ is considered relevant to a harmful protein $p^i_\text{H}$ if $\text{dis}(p^i_\text{H}, p^j_\text{B}) \leq \tau$, where $\tau$ is a predefined hop threshold. This selection criterion enables us to identify benign proteins that are structurally close to harmful proteins, capturing potential functional or biological similarities. 

At the embedding level, we leverage the TransE algorithm to learn low-dimensional representations of nodes in the PSKG. Each node $v_i \in V^{'}$ is embedded into a vector space as $e_i \in \mathbb{R}^d$, where $d$ is the embedding dimension. 
The TransE algorithm optimizes the following objective function:
\begin{equation}
L\!=\!\!\!\!\!\!\!\!\!\!\sum_{(v_i,r,v_j) \in E}\!\!\!\!\!\!\![\Vert e_i+r-e_j\Vert ^2\!+\!\!\!\!\!\!\!\!\!\sum_{(v^{'}_{i},r,v^{'}_{j}) \notin E}\!\!\!\!\!\!\!\!\text{max}(0,\eta-\Vert e^{'}_i+r-e^{'}_j\Vert ^2)], \label{eq:7}
\end{equation}
where $r \in \mathbb{R}^d$ represents the embedding of the edge relation, and $\eta$ is a margin hyperparameter. This formulation ensures that embeddings reflect the structural relationships in the graph, such that semantically similar nodes are embedded closer together. 
By combining structural proximity and embedding-based similarity, we identify benign proteins $p^j_\text{B}$ that are closely related to harmful proteins $p^i_\text{H}$ in both the graph structure and the embedding space. For each harmful protein $p^i_\text{H}$, we select the top-$M$ benign proteins based on a combined similarity score:
\begin{equation}
s(p^i_\text{H}, p^j_\text{B}) = \mu \cdot \frac{1}{\text{dis}(p^i_\text{H}, p^j_\text{B})} + (1 - \mu) \cdot \cos(e_{p^i_\text{H}}, e_{p^j_\text{B}}), \label{eq:8}
\end{equation}
where $\cos(\cdot, \cdot)$ denotes the cosine similarity between embeddings, and $\mu \in [0, 1]$ is a weighting hyper-parameter.

Using these selected pairs of harmful and benign proteins, we construct preference pairs for fine-tuning the PLM. Each preference pair $(p^j_\text{B}, p^i_\text{H})$ encodes the notion that the model should prefer generating sequences similar to the benign protein $p^j_\text{B}$ over the harmful protein $p^i_\text{H}$. The fine-tuning process is guided by Direct Preference Optimization, which optimizes the following objective:
\begin{equation}
L_{\text{KPO}}\!\! =\! - \log\! \sigma\!\left( \varphi\!\ \cdot\!\! \left[ \log P_\theta(p^j_\text{B} | x)\! - \! \log P_\theta(p^i_\text{H} | x) \right] \right), \label{eq:9}
\end{equation}
where $P_\theta(\cdot | x)$ is the probability assigned by the PLM to a sequence given an input prompt or context $x$, $\sigma(\cdot)$ is the sigmoid function, and $\varphi$ is a scaling factor.
By fine-tuning the PLM with these preference pairs, the model learns to distinguish subtle differences between harmful and benign proteins. This process integrates the rich biological information in the PSKG into the generative model, significantly enhancing its ability to generate safe and biologically meaningful protein sequences. 

\newcommand{\myfont}{\fontsize{9.2pt}{\baselineskip}\selectfont}

\begin{table*}[h]\myfont
\begin{center}
\begin{minipage}{\textwidth}
\caption{Performance of KPO on the top of different base models.}
\label{tab1}
\begin{tabularx}{\textwidth}{@{\extracolsep{\fill}}p{2.2cm}cccccccccc@{\extracolsep{\fill}}}
\toprule%
\multirow{2}{*}{Models} & \multicolumn{5}{@{}c@{}}{Safety Evaluation Metric} & \multicolumn{4}{@{}c@{}}{Functional Evaluation Metric} \\
\cmidrule(lr){2-6}\cmidrule(lr){7-10}%
& BLAST$\mathord{\downarrow}$ & MMseq2$\mathord{\downarrow}$ & Pfam\_D$\mathord{\downarrow}$ & Pfam\_E$\mathord{\downarrow}$ & ToxinPred3$\mathord{\downarrow}$ & GB1$\mathord{\uparrow}$ & PhoQ$\mathord{\uparrow}$ & UBC9$\mathord{\uparrow}$ & GFP$\mathord{\uparrow}$ \\
\midrule
ProtGPT2  & 0.269 & 0.325 & 0.2933 & 0.2701 & 0.07 & 0.030 & 0.015 & \textbf{0.138} & 1.526\\
ProtGPT2\textbf{+KPO} & \textbf{0.138} & \textbf{0.149} & \textbf{0.2613} & \textbf{0.1819} & \textbf{0.024} & \textbf{0.041} & \textbf{0.315} & 0.129 & \textbf{2.204}\\ \midrule
Progen2 & 0.155 & 0.170 & 0.1079 & 0.2630 & 0.029 & \textbf{0.144} & \textbf{0.027} & 0.068 & \textbf{1.683} \\
Progen2\textbf{+KPO} & \textbf{0.128} & \textbf{0.117} & \textbf{0.0922}
 & \textbf{0.0645} & \textbf{0.007} & 0.024 & 0.017 & \textbf{0.231} & 1.562 \\ \midrule
 InstructProtein & 0.410 & 0.285 & 0.1662 & 0.0835 & 0.031
 & 0.030 & 0.016 & \textbf{0.459} & 1.983\\
InstructProtein\textbf{+KPO} & \textbf{0.086} & \textbf{0.079} & \textbf{0.1189} & \textbf{0.0385} & \textbf{0.003} & \textbf{0.191} & \textbf{0.814} & 0.451 & \textbf{2.319}\\
\bottomrule
\end{tabularx}
\end{minipage}
\end{center}
\end{table*}

\section{Experimental Results}

\subsection{Experimental Settings}

\paragraph{\textbf{Safety Evaluation Metrics}} 
In evaluating the harmfulness of generated protein sequences, we employed a comprehensive approach that incorporated sequence similarity analysis, functional domain characterization, and toxicity prediction. This allowed us to rigorously assess the potential risks of the generated proteins, ensuring a robust evaluation of their safety and biological relevance.

For sequence similarity analysis, we utilized BLAST~\cite{madden2013blast} and MMseqs2~\cite{steinegger2017mmseqs2}, two widely adopted tools in bioinformatics. To quantify this similarity, we computed the average alignment score for each generated protein sequence against a test dataset of harmful proteins, normalizing the score by the average sequence length. This normalization allowed for a fair comparison across proteins of varying lengths and ensured that the sequence length did not disproportionately influence the similarity score. 

In the functional domain analysis, we turned to Pfam~\cite{finn2014pfam}, a database that provides Hidden Markov Models (HMMs)~\cite{eddy1996hidden} for identifying protein domains. Each generated protein sequence and the harmful protein test dataset were analyzed for functional domain content. 
For domains identified in both the generated sequences and harmful proteins, we employed two distinct strategies to evaluate the significance of these domain matches. The first strategy used dynamic E-value thresholds, where we compared the E-values of functional domains in the generated proteins to those in the harmful protein database. This approach accounts for the unique statistical
characteristics of each domain\cite{mistry2021pfam}. The second strategy employed a fixed E-value threshold of 0.001\cite{finn2014pfam}, a commonly accepted cutoff for statistically significant domain matches. Lower E-values indicate a high likelihood of functional similarity to harmful proteins, while higher E-values suggest weaker or no significant functional relationship. See Appendix \ref{ap:ex} for specific reasons for this setting.

In addition, we incorporated ToxinPred3, a machine learning classifier to predict the toxicity of the generated proteins~\cite{rathore2024toxinpred}. ToxinPred3 outputs a probability score indicating the likelihood of a protein being toxic. Proteins with toxicity scores above a predefined threshold were classified as toxic. The number of toxic predictions before and after fine-tuning was compared to assess the success of our approach in reducing the generation of potentially harmful proteins.


\paragraph{\textbf{Functional Evaluation Metrics}} 
To assess if the functional capabilities of the pre-trained model remained intact after fine-tuning using the KPO method, we employed a set of well-established protein datasets: GB1~\cite{wu2019machine}, PhoQ~\cite{podgornaia2015pervasive}, UBC9~\cite{knipscheer2008ubc9}, and GFP~\cite{zimmer2002green}. For each of these datasets, we compared the performance of the pre-trained model and the fine-tuned model by calculating the generation probability score for each mutation. We computed the scores for the top 96 mutations with the highest probabilities generated by both the pre-trained and fine-tuned models. To ensure that fine-tuning did not compromise the model’s ability to generate functionally relevant mutations, we calculated the average fitness of these top 96 mutations. Specific implementation details are provided in the Appendix \ref{ap:ex}.

\subsection{Main Results}
\paragraph{\textbf{Performance Comparison}}
The performance of our KPO method, applied to three different base models published in top-tier journals or conferences—ProtGPT2~\cite{ferruz2022protgpt2}, Progen2~\cite{nijkamp2023progen2} and InstructProtein~\cite{wang2023instructprotein}—was evaluated across both safety and functional metrics. The results demonstrate that KPO significantly enhances the safety of generated protein sequences, while preserving or even improving the functional capabilities of the models. This balance between safety and functionality highlights the efficacy of KPO in optimizing protein sequence generation. Our code is available at https://github.com/HICAI-ZJU/KPO.

\begin{figure}[t]
    \centering
    \includegraphics[width=\columnwidth]{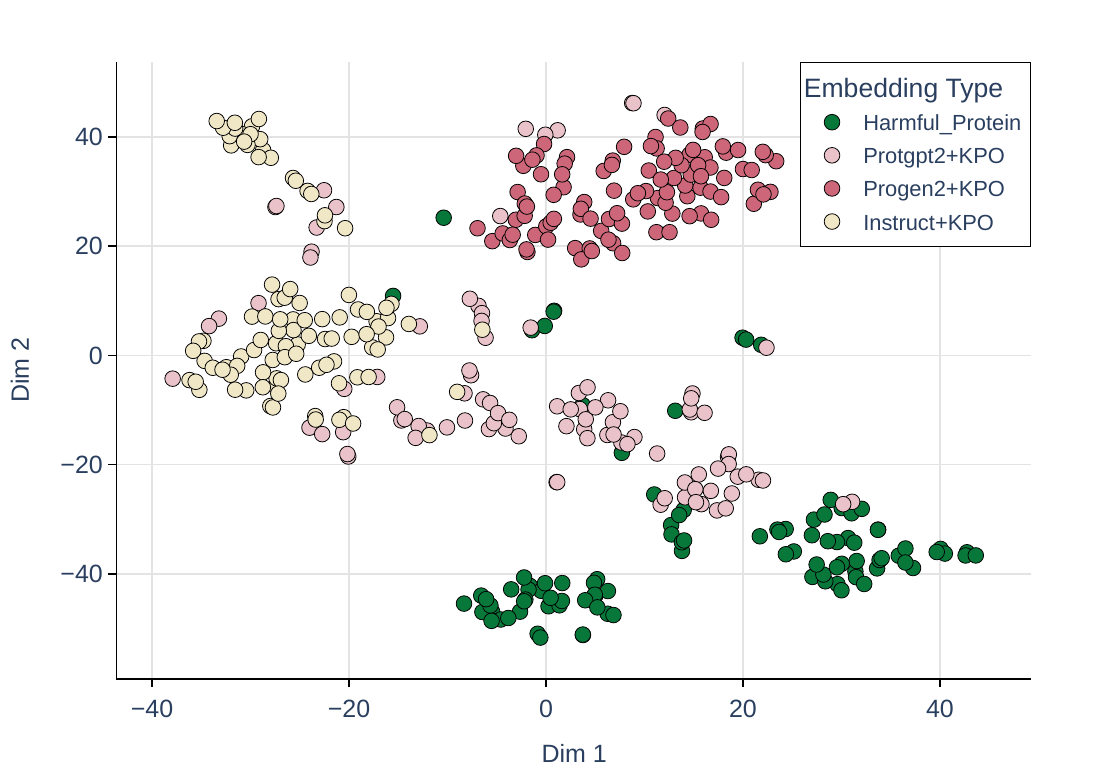}
    \caption{
    The embeddings of proteins generated by the fine-tuned PLM exhibit a clear separation from those of harmful proteins, demonstrating that the KPO framework effectively steers the PLM away from harmful regions in the latent space.}
    \label{fig:embedding}
\end{figure}

From Table \ref{tab1}, we can observe that the proposed KPO method consistently reduced the risk of generating harmful proteins across multiple metrics. Additionally, the functional domain-based harmfulness assessment using Pfam also showed improvements.  The ToxinPred3 toxicity classifier also showed a sharp reduction in predicted toxicity. 
One of the reasons for the observed functional improvements after applying KPO is that the fine-tuning process effectively guides the model away from harmful sequence spaces, which can be biologically unproductive, and toward regions of the sequence space that are more likely to produce functional and high-adaptability proteins. By minimizing the generation of harmful proteins, the model avoids the regions that might be associated with harmful or suboptimal structural configurations. As a result, the model can explore the most promising areas of the protein sequence landscape, where the generated proteins exhibit higher fitness and functional relevance. This allows the protein model to focus better on generating sequences that are not only safe but also biologically advantageous, improving its ability to optimize for functions without compromising safety.

\begin{figure*}[t]
    \centering
    \includegraphics[width=2\columnwidth]{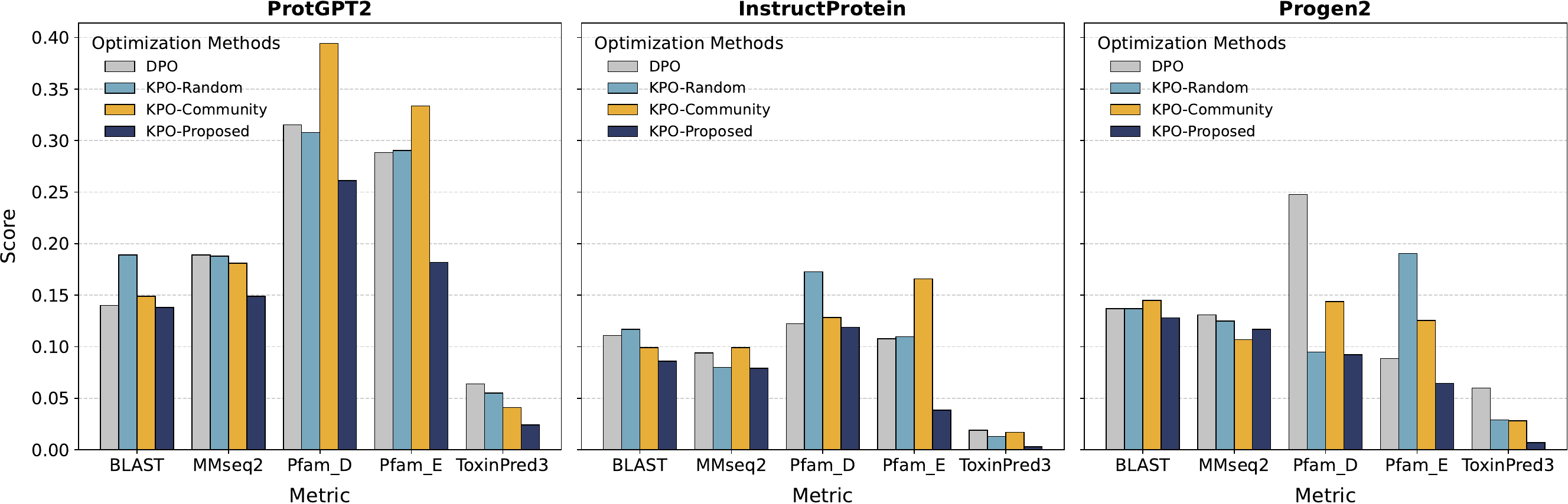}
    \caption{Ablation study results for three PLMs: ProtGPT2 (left), InstructProtein (middle), and ProGen2 (right). Each plot compares the performance of different optimization methods across five safety evaluation metrics. 
    }
    \label{fig:Ablation}
\end{figure*} 

\paragraph{\textbf{Embedding Result Analysis}}
To further investigate the impact of KPO on the generated protein sequences, we analyzed their embeddings using the ESM-2 ~\cite{verkuil2022language} model. This analysis aimed to visualize how the fine-tuned models' generated sequences compared to known harmful protein sequences in the embedding space. Specifically, we selected 100 protein sequences from each of the following categories: harmful proteins, and proteins generated by the KPO fine-tuned models (ProtGPT2, Progen2, and InstructProtein). These sequences were then processed through the ESM-2 model to obtain their respective embeddings, which were subsequently visualized using the t-SNE ~\cite{van2008visualizing} algorithm for dimensionality reduction.

Figure \ref{fig:embedding} demonstrates a significant divergence between the harmful protein embeddings and those of the generated proteins from the KPO-fine-tuned models. The sequences generated by the KPO-optimized ProtGPT2, Progen2, and InstructProtein models form distinct clusters that are well-separated from the cluster representing the harmful protein embeddings. This separation indicates that the fine-tuning process effectively steered the models away from generating sequences that share structural or functional similarities with known harmful proteins. This shift in embedding space aligns with the improvements observed in safety evaluation metrics, such as BLAST, MMseqs2, and ToxinPred3, further validating the effectiveness of the fine-tuning process.

\newcommand{\myf}{\fontsize{10.2pt}{\baselineskip}\selectfont}

\begin{table*}[t]\myf
\centering
\begin{minipage}{\textwidth}
\caption{Performance under different $\alpha, \beta$ combinations.}
\label{tab:alpha_beta_performance}
\begin{tabularx}{\textwidth}{@{\extracolsep{\fill}}p{1.2cm}cccccccccc@{\extracolsep{\fill}}}
\toprule
$\alpha, \beta$ & BLAST$\mathord{\downarrow}$ & MMseq2$\mathord{\downarrow}$ & Pfam\_D$\mathord{\downarrow}$ & Pfam\_E$\mathord{\downarrow}$ & ToxinPred3$\mathord{\downarrow}$ & GB1$\mathord{\uparrow}$ & PhoQ$\mathord{\uparrow}$ & UBC9$\mathord{\uparrow}$ & GFP$\mathord{\uparrow}$ \\
\midrule
0.1, 0.9 & 0.111 & 0.125 & 0.294 & 0.245 & 0.017 & 0.062 & 0.025 & 0.137 & 1.68 \\
0.3, 0.7 & 0.107 & 0.102 & 0.318 & 0.241 & 0.017 & 0.033 & 0.001 & \textbf{0.169} & 2.12 \\
0.5, 0.5 & 0.138 & 0.149 & 0.261 & \textbf{0.182} & 0.024 & 0.041 & 0.315 & 0.129 & \textbf{2.20} \\
0.7, 0.3 & \textbf{0.100} & \textbf{0.100} & 0.351 & 0.248 & \textbf{0.010} & \textbf{0.098} & 0.090 & 0.099 & 2.13 \\
0.9, 0.1 & 0.148 & 0.145 & \textbf{0.181} & 0.262 & 0.014 & 0.046 & \textbf{0.907} & 0.100 & 1.66 \\
\bottomrule
\end{tabularx}
\end{minipage}
\end{table*}

\begin{table*}[t]\myf
\centering
\begin{minipage}{\textwidth}
\caption{Performance under different $\gamma$:$\sigma$ combinations.}
\label{tab:gamma_sigma_performance}
\begin{tabularx}{\textwidth}{@{\extracolsep{\fill}}p{1.2cm}cccccccccc@{\extracolsep{\fill}}}
\toprule
$\gamma$:$\sigma$ & BLAST$\mathord{\downarrow}$ & MMseq2$\mathord{\downarrow}$ & Pfam\_D$\mathord{\downarrow}$ & Pfam\_E$\mathord{\downarrow}$ & ToxinPred3$\mathord{\downarrow}$ & GB1$\mathord{\uparrow}$ & PhoQ$\mathord{\uparrow}$ & UBC9$\mathord{\uparrow}$ & GFP$\mathord{\uparrow}$ \\
\midrule
1:1 & \textbf{0.108} & 0.154 & \textbf{0.171} & 0.228 & \textbf{0.002} & \textbf{0.067} & 0.116 & 0.138 & 2.16 \\
1:2 & 0.138 & 0.149 & 0.261 & \textbf{0.182} & 0.024 & 0.041 & 0.315 & 0.129 & 2.20 \\
1:3 & 0.119 & \textbf{0.111} & 0.194 & 0.263 & 0.011 & 0.026 & \textbf{0.916} & 0.145 & \textbf{2.30} \\
2:1 & 0.137 & 0.125 & 0.179 & 0.232 & 0.026 & 0.049 & 0.111 & 0.114 & 1.59 \\
3:1 & 0.153 & 0.153 & 0.273 & 0.274 & 0.028 & 0.052 & 0.001 & \textbf{0.161} & 1.54 \\
\bottomrule
\end{tabularx}
\end{minipage}
\end{table*}

\subsection{Ablation Study}

To validate the effectiveness of our PSKG and the proposed graph pruning strategy, we conducted an extensive ablation study. This study compared KPO with three alternative methods: DPO, KPO-random, and KPO-community. Each alternative serves as a variation in the use of PSKG or sampling strategies during the fine-tuning process, enabling a comprehensive evaluation of KPO's core components and their contributions.

For the DPO baseline, the PSKG was not utilized. Instead, benign proteins were randomly sampled from the Uniprot database to construct the preference dataset, and the standard DPO method was applied for fine-tuning. This evaluates the benefits of incorporating structured domain knowledge into the fine-tuning process. For the KPO-random method, the PSKG was utilized but reduced in scale by randomly pruning nodes and edges. The reduced graph was then used to select benign proteins for fine-tuning with DPO. This method tests the effectiveness of random graph reduction compared to structured pruning. In contrast, the KPO-community method used a community detection algorithm to reduce the scale of the PSKG. By clustering nodes based on their structural relationships, this approach preserved local community structures within the graph while reducing its overall size. Benign proteins were then selected from these communities for DPO fine-tuning. This method evaluates the impact of structural preservation during graph reduction on model performance. The experimental results, summarized in Figure \ref{fig:Ablation}, show that the KPO method consistently outperformed the alternative methods across nearly all safety evaluation metrics, including BLAST, MMseqs2, Pfam\_D, Pfam\_E, and ToxinPred3.The results also highlight the effectiveness of the graph pruning strategy used in KPO. Both KPO-random and KPO-community served as benchmarks to test the validity of our pruning method. The ablation study confirms that the proposed graph pruning algorithm in KPO is not only effective in reducing the graph’s scale but also preserves essential biological relationships.

To further assess the robustness of the KPO framework, we conducted additional experiments and sensitivity analyses using ProtGPT2 under a range of hyperparameter configurations. The hyperparameters $\alpha$, $\beta$, $\gamma$, $\delta$, $Q$, and $K$ are central to our pruning strategy, and we provide a comprehensive assessment of their influence on model performance. The parameters $\alpha$ and $\beta$ control the trade-off between the connectivity of high-scoring GO nodes ($\alpha$) and the degree centrality of benign protein nodes ($\beta$). As reported in Table~\ref{tab:alpha_beta_performance}, the model demonstrates stable performance across various ($\alpha$, $\beta$) combinations, with safety evaluation metrics (e.g., BLAST, MMseq2) exhibiting only minor variations. Similarly, $\gamma$ and $\delta$ balance the harmful-benign bridging degree and the neighborhood breadth, respectively. Table~\ref{tab:gamma_sigma_performance} shows that changes in the $\gamma$:$\delta$ ratio do not substantially impact model performance. In particular, the configuration $\gamma$:$\delta$ = 1:1 yields optimal results for most evaluation metrics, especially in toxicity prediction (ToxinPred3), where the harmfulness score is notably reduced (0.002). These results indicate that the pruning strategy is robust to variations in the balance between structural relevance and functional connectivity. Furthermore, we performed comprehensive sensitivity analyses for $Q$ and $K$, with detailed results provided in Appendix~\ref{ap:Sensitivity}. These findings collectively demonstrate that the KPO framework is robust to variations in key hyperparameters, further supporting the reliability of our proposed approach.

\section{Conclusion and Future Work}
In this study, we proposed Knowledge-guided Preference Optimization (KPO), a method for fine-tuning protein language models to ensure the safety and functionality of generated sequences. By constructing a comprehensive dataset of Protein Safety Knowledge Graph and integrating it with pruning strategies and reinforcement learning, KPO reduces the risk of generating harmful proteins while maintaining or enhancing functional performance. This work provides a framework for incorporating safety knowledge into PLMs, enabling responsible applications in protein engineering. Future work will focus on extending KPO by integrating structural safety constraints, scaling to larger PLMs, and dynamically updating the PSKG with new safety insights, ensuring adaptability to evolving biological challenges.

\section*{Acknowledgements}
This work was supported by the “Pioneer” and “Leading Goose”
R\&D Program of Zhejiang (Grant No. 2025C01097), Hangzhou West Lake Pearl Project Leading Innovative Youth Team Project (TD2023017) and the Ningbo Natural Science Foundation (2024J020).


\section*{Limitations}

While the KPO framework demonstrates promising results in aligning PLMs with safety constraints, several limitations should be addressed in future work. First, 
the fine-tuning process in this paper focuses primarily on sequence-level safety constraints. Structural-level constraints, such as ensuring generated proteins avoid toxic 3D conformations, are not directly incorporated and rely on downstream evaluation tools. In addition, while reinforcement learning enables nuanced optimization, the computational overhead for training large PLMs remains significant, limiting its scalability to larger datasets or more complex safety objectives. Addressing these limitations will be critical for advancing the robustness and applicability of KPO in real-world scenarios, such as therapeutic development and environmental biotechnology.



\bibliography{custom}

\begin{thebibliography}{43}
\providecommand{\natexlab}[1]{#1}

\bibitem[{Bai et~al.(2022)Bai, Jones, Ndousse, Askell, Chen, DasSarma, Drain, Fort, Ganguli, Henighan et~al.}]{bai2022training}
Yuntao Bai, Andy Jones, Kamal Ndousse, Amanda Askell, Anna Chen, Nova DasSarma, Dawn Drain, Stanislav Fort, Deep Ganguli, Tom Henighan, et~al. 2022.
\newblock Training a helpful and harmless assistant with reinforcement learning from human feedback.
\newblock \emph{arXiv preprint arXiv:2204.05862}.

\bibitem[{Bourtoule et~al.(2021)Bourtoule, Chandrasekaran, Choquette-Choo, Jia, Travers, Zhang, Lie, and Papernot}]{bourtoule2021machine}
Lucas Bourtoule, Varun Chandrasekaran, Christopher~A Choquette-Choo, Hengrui Jia, Adelin Travers, Baiwu Zhang, David Lie, and Nicolas Papernot. 2021.
\newblock Machine unlearning.
\newblock In \emph{2021 IEEE Symposium on Security and Privacy (SP)}, pages 141--159. IEEE.

\bibitem[{Chen(2023)}]{chen2023large}
Huajun Chen. 2023.
\newblock Large knowledge model: Perspectives and challenges.
\newblock \emph{arXiv preprint arXiv:2312.02706}.

\bibitem[{Eddy(1996)}]{eddy1996hidden}
Sean~R Eddy. 1996.
\newblock Hidden markov models.
\newblock \emph{Current opinion in structural biology}, 6(3):361--365.

\bibitem[{Ferruz et~al.(2022)Ferruz, Schmidt, and H{\"o}cker}]{ferruz2022protgpt2}
Noelia Ferruz, Steffen Schmidt, and Birte H{\"o}cker. 2022.
\newblock Protgpt2 is a deep unsupervised language model for protein design.
\newblock \emph{Nature communications}, 13(1):4348.

\bibitem[{Finn et~al.(2014)Finn, Bateman, Clements, Coggill, Eberhardt, Eddy, Heger, Hetherington, Holm, Mistry et~al.}]{finn2014pfam}
Robert~D Finn, Alex Bateman, Jody Clements, Penelope Coggill, Ruth~Y Eberhardt, Sean~R Eddy, Andreas Heger, Kirstie Hetherington, Liisa Holm, Jaina Mistry, et~al. 2014.
\newblock Pfam: the protein families database.
\newblock \emph{Nucleic acids research}, 42(D1):D222--D230.

\bibitem[{He et~al.(2023)He, He, Guan, Zhao, Chen, Zhu, Chen, Li, and Yao}]{he2023novo}
HaoHuai He, Bing He, Lei Guan, Yu~Zhao, Guanxing Chen, Qingge Zhu, Calvin Yu-Chian Chen, Ting Li, and Jianhua Yao. 2023.
\newblock De novo generation of antibody cdrh3 with a pre-trained generative large language model.
\newblock \emph{bioRxiv}, pages 2023--10.

\bibitem[{Ishibashi and Shimodaira(2023)}]{ishibashi2023knowledge}
Yoichi Ishibashi and Hidetoshi Shimodaira. 2023.
\newblock Knowledge sanitization of large language models.
\newblock \emph{arXiv preprint arXiv:2309.11852}.

\bibitem[{Jang et~al.(2022)Jang, Yoon, Yang, Cha, Lee, Logeswaran, and Seo}]{jang2022knowledge}
Joel Jang, Dongkeun Yoon, Sohee Yang, Sungmin Cha, Moontae Lee, Lajanugen Logeswaran, and Minjoon Seo. 2022.
\newblock Knowledge unlearning for mitigating privacy risks in language models.
\newblock \emph{arXiv preprint arXiv:2210.01504}.

\bibitem[{Jumper et~al.(2021)Jumper, Evans, Pritzel, Green, Figurnov, Ronneberger, Tunyasuvunakool, Bates, {\v{Z}}{\'\i}dek, Potapenko et~al.}]{jumper2021highly}
John Jumper, Richard Evans, Alexander Pritzel, Tim Green, Michael Figurnov, Olaf Ronneberger, Kathryn Tunyasuvunakool, Russ Bates, Augustin {\v{Z}}{\'\i}dek, Anna Potapenko, et~al. 2021.
\newblock Highly accurate protein structure prediction with alphafold.
\newblock \emph{nature}, 596(7873):583--589.

\bibitem[{Kassem et~al.(2023)Kassem, Mahmoud, and Saad}]{kassem2023preserving}
Aly Kassem, Omar Mahmoud, and Sherif Saad. 2023.
\newblock Preserving privacy through dememorization: An unlearning technique for mitigating memorization risks in language models.
\newblock In \emph{Proceedings of the 2023 Conference on Empirical Methods in Natural Language Processing}, pages 4360--4379.

\bibitem[{Knipscheer et~al.(2008)Knipscheer, Flotho, Klug, Olsen, van Dijk, Fish, Johnson, Mann, Sixma, and Pichler}]{knipscheer2008ubc9}
Puck Knipscheer, Annette Flotho, Helene Klug, Jesper~V Olsen, Willem~J van Dijk, Alexander Fish, Erica~S Johnson, Matthias Mann, Titia~K Sixma, and Andrea Pichler. 2008.
\newblock Ubc9 sumoylation regulates sumo target discrimination.
\newblock \emph{Molecular cell}, 31(3):371--382.

\bibitem[{Li et~al.(2024)Li, Zhou, Tan, and Hong}]{li2024unlearning}
Mingchen Li, Bingxin Zhou, Yang Tan, and Liang Hong. 2024.
\newblock Unlearning virus knowledge toward safe and responsible mutation effect predictions.
\newblock \emph{bioRxiv}, pages 2024--10.

\bibitem[{Lin et~al.(2022)Lin, Akin, Rao, Hie, Zhu, Lu, dos Santos~Costa, Fazel-Zarandi, Sercu, Candido et~al.}]{lin2022language}
Zeming Lin, Halil Akin, Roshan Rao, Brian Hie, Zhongkai Zhu, Wenting Lu, Allan dos Santos~Costa, Maryam Fazel-Zarandi, Tom Sercu, Sal Candido, et~al. 2022.
\newblock Language models of protein sequences at the scale of evolution enable accurate structure prediction.
\newblock \emph{BioRxiv}, 2022:500902.

\bibitem[{Lu et~al.(2022)Lu, Welleck, Hessel, Jiang, Qin, West, Ammanabrolu, and Choi}]{lu2022quark}
Ximing Lu, Sean Welleck, Jack Hessel, Liwei Jiang, Lianhui Qin, Peter West, Prithviraj Ammanabrolu, and Yejin Choi. 2022.
\newblock Quark: Controllable text generation with reinforced unlearning.
\newblock \emph{Advances in neural information processing systems}, 35:27591--27609.

\bibitem[{Madani et~al.(2023)Madani, Krause, Greene, Subramanian, Mohr, Holton, Olmos, Xiong, Sun, Socher et~al.}]{madani2023large}
Ali Madani, Ben Krause, Eric~R Greene, Subu Subramanian, Benjamin~P Mohr, James~M Holton, Jose~Luis Olmos, Caiming Xiong, Zachary~Z Sun, Richard Socher, et~al. 2023.
\newblock Large language models generate functional protein sequences across diverse families.
\newblock \emph{Nature Biotechnology}, 41(8):1099--1106.

\bibitem[{Madden(2013)}]{madden2013blast}
Thomas Madden. 2013.
\newblock The blast sequence analysis tool.
\newblock \emph{The NCBI handbook}, 2(5):425--436.

\bibitem[{Mirdita et~al.(2022)Mirdita, Sch{\"u}tze, Moriwaki, Heo, Ovchinnikov, and Steinegger}]{mirdita2022colabfold}
Milot Mirdita, Konstantin Sch{\"u}tze, Yoshitaka Moriwaki, Lim Heo, Sergey Ovchinnikov, and Martin Steinegger. 2022.
\newblock Colabfold: making protein folding accessible to all.
\newblock \emph{Nature methods}, 19(6):679--682.

\bibitem[{Mistry et~al.(2021)Mistry, Chuguransky, Williams, Qureshi, Salazar, Sonnhammer, Tosatto, Paladin, Raj, Richardson et~al.}]{mistry2021pfam}
Jaina Mistry, Sara Chuguransky, Lowri Williams, Matloob Qureshi, Gustavo~A Salazar, Erik~LL Sonnhammer, Silvio~CE Tosatto, Lisanna Paladin, Shriya Raj, Lorna~J Richardson, et~al. 2021.
\newblock Pfam: The protein families database in 2021.
\newblock \emph{Nucleic acids research}, 49(D1):D412--D419.

\bibitem[{Nijkamp et~al.(2023)Nijkamp, Ruffolo, Weinstein, Naik, and Madani}]{nijkamp2023progen2}
Erik Nijkamp, Jeffrey~A Ruffolo, Eli~N Weinstein, Nikhil Naik, and Ali Madani. 2023.
\newblock Progen2: exploring the boundaries of protein language models.
\newblock \emph{Cell systems}, 14(11):968--978.

\bibitem[{Ouyang et~al.(2022)Ouyang, Wu, Jiang, Almeida, Wainwright, Mishkin, Zhang, Agarwal, Slama, Ray et~al.}]{ouyang2022training}
Long Ouyang, Jeffrey Wu, Xu~Jiang, Diogo Almeida, Carroll Wainwright, Pamela Mishkin, Chong Zhang, Sandhini Agarwal, Katarina Slama, Alex Ray, et~al. 2022.
\newblock Training language models to follow instructions with human feedback.
\newblock \emph{Advances in neural information processing systems}, 35:27730--27744.

\bibitem[{Pawelczyk et~al.(2023)Pawelczyk, Neel, and Lakkaraju}]{pawelczyk2023context}
Martin Pawelczyk, Seth Neel, and Himabindu Lakkaraju. 2023.
\newblock In-context unlearning: Language models as few shot unlearners.
\newblock \emph{arXiv preprint arXiv:2310.07579}.

\bibitem[{Podgornaia and Laub(2015)}]{podgornaia2015pervasive}
Anna~I Podgornaia and Michael~T Laub. 2015.
\newblock Pervasive degeneracy and epistasis in a protein-protein interface.
\newblock \emph{Science}, 347(6222):673--677.

\bibitem[{Pokharel et~al.(2022)Pokharel, Pratyush, Heinzinger, Newman, and Kc}]{pokharel2022improving}
Suresh Pokharel, Pawel Pratyush, Michael Heinzinger, Robert~H Newman, and Dukka~B Kc. 2022.
\newblock Improving protein succinylation sites prediction using embeddings from protein language model.
\newblock \emph{Scientific reports}, 12(1):16933.

\bibitem[{Rafailov et~al.(2024)Rafailov, Sharma, Mitchell, Manning, Ermon, and Finn}]{rafailov2024direct}
Rafael Rafailov, Archit Sharma, Eric Mitchell, Christopher~D Manning, Stefano Ermon, and Chelsea Finn. 2024.
\newblock Direct preference optimization: Your language model is secretly a reward model.
\newblock \emph{Advances in Neural Information Processing Systems}, 36.

\bibitem[{Rao et~al.(2021)Rao, Liu, Verkuil, Meier, Canny, Abbeel, Sercu, and Rives}]{rao2021msa}
Roshan~M Rao, Jason Liu, Robert Verkuil, Joshua Meier, John Canny, Pieter Abbeel, Tom Sercu, and Alexander Rives. 2021.
\newblock Msa transformer.
\newblock In \emph{International Conference on Machine Learning}, pages 8844--8856. PMLR.

\bibitem[{Rathore et~al.(2024)Rathore, Choudhury, Arora, Tijare, and Raghava}]{rathore2024toxinpred}
Anand~Singh Rathore, Shubham Choudhury, Akanksha Arora, Purva Tijare, and Gajendra~PS Raghava. 2024.
\newblock Toxinpred 3.0: An improved method for predicting the toxicity of peptides.
\newblock \emph{Computers in Biology and Medicine}, 179:108926.

\bibitem[{Rives et~al.(2021)Rives, Meier, Sercu, Goyal, Lin, Liu, Guo, Ott, Zitnick, Ma et~al.}]{rives2021biological}
Alexander Rives, Joshua Meier, Tom Sercu, Siddharth Goyal, Zeming Lin, Jason Liu, Demi Guo, Myle Ott, C~Lawrence Zitnick, Jerry Ma, et~al. 2021.
\newblock Biological structure and function emerge from scaling unsupervised learning to 250 million protein sequences.
\newblock \emph{Proceedings of the National Academy of Sciences}, 118(15):e2016239118.

\bibitem[{Steinegger and S{\"o}ding(2017)}]{steinegger2017mmseqs2}
Martin Steinegger and Johannes S{\"o}ding. 2017.
\newblock Mmseqs2 enables sensitive protein sequence searching for the analysis of massive data sets.
\newblock \emph{Nature biotechnology}, 35(11):1026--1028.

\bibitem[{Van~der Maaten and Hinton(2008)}]{van2008visualizing}
Laurens Van~der Maaten and Geoffrey Hinton. 2008.
\newblock Visualizing data using t-sne.
\newblock \emph{Journal of machine learning research}, 9(11).

\bibitem[{Verkuil et~al.(2022)Verkuil, Kabeli, Du, Wicky, Milles, Dauparas, Baker, Ovchinnikov, Sercu, and Rives}]{verkuil2022language}
Robert Verkuil, Ori Kabeli, Yilun Du, Basile~IM Wicky, Lukas~F Milles, Justas Dauparas, David Baker, Sergey Ovchinnikov, Tom Sercu, and Alexander Rives. 2022.
\newblock Language models generalize beyond natural proteins.
\newblock \emph{bioRxiv}, pages 2022--12.

\bibitem[{Vig et~al.(2020)Vig, Madani, Varshney, Xiong, Socher, and Rajani}]{vig2020bertology}
Jesse Vig, Ali Madani, Lav~R Varshney, Caiming Xiong, Richard Socher, and Nazneen~Fatema Rajani. 2020.
\newblock Bertology meets biology: Interpreting attention in protein language models.
\newblock \emph{arXiv preprint arXiv:2006.15222}.

\bibitem[{von Werra et~al.(2020)von Werra, Belkada, Tunstall, Beeching, Thrush, Lambert, Huang, Rasul, and Gallouédec}]{vonwerra2022trl}
Leandro von Werra, Younes Belkada, Lewis Tunstall, Edward Beeching, Tristan Thrush, Nathan Lambert, Shengyi Huang, Kashif Rasul, and Quentin Gallouédec. 2020.
\newblock Trl: Transformer reinforcement learning.
\newblock \url{https://github.com/huggingface/trl}.

\bibitem[{Wang et~al.(2024)Wang, Patsenker, Li, Kluger, and Kleinstein}]{wang2024supervised}
Meng Wang, Jonathan Patsenker, Henry Li, Yuval Kluger, and Steven~H Kleinstein. 2024.
\newblock Supervised fine-tuning of pre-trained antibody language models improves antigen specificity prediction.
\newblock \emph{bioRxiv}.

\bibitem[{Wang et~al.(2023)Wang, Zhang, Ding, Qin, Zhuang, Li, and Chen}]{wang2023instructprotein}
Zeyuan Wang, Qiang Zhang, Keyan Ding, Ming Qin, Xiang Zhuang, Xiaotong Li, and Huajun Chen. 2023.
\newblock Instructprotein: Aligning human and protein language via knowledge instruction.
\newblock \emph{arXiv preprint arXiv:2310.03269}.

\bibitem[{Wang et~al.(2022)Wang, Combs, Brand, Calvo, Xu, Price, Golovach, Salawu, Wise, Ponnapalli et~al.}]{wang2022lm}
Zichen Wang, Steven~A Combs, Ryan Brand, Miguel~Romero Calvo, Panpan Xu, George Price, Nataliya Golovach, Emmanuel~O Salawu, Colby~J Wise, Sri~Priya Ponnapalli, et~al. 2022.
\newblock Lm-gvp: an extensible sequence and structure informed deep learning framework for protein property prediction.
\newblock \emph{Scientific reports}, 12(1):6832.

\bibitem[{Wu et~al.(2019)Wu, Kan, Lewis, Wittmann, and Arnold}]{wu2019machine}
Zachary Wu, SB~Jennifer Kan, Russell~D Lewis, Bruce~J Wittmann, and Frances~H Arnold. 2019.
\newblock Machine learning-assisted directed protein evolution with combinatorial libraries.
\newblock \emph{Proceedings of the National Academy of Sciences}, 116(18):8852--8858.

\bibitem[{Yang et~al.(2023)Yang, Klein, Celikyilmaz, Peng, and Tian}]{yang2023rlcd}
Kevin Yang, Dan Klein, Asli Celikyilmaz, Nanyun Peng, and Yuandong Tian. 2023.
\newblock Rlcd: Reinforcement learning from contrast distillation for language model alignment.
\newblock \emph{arXiv preprint arXiv:2307.12950}.

\bibitem[{Yuan et~al.(2023)Yuan, Yuan, Tan, Wang, Huang, and Huang}]{yuan2023rrhf}
Zheng Yuan, Hongyi Yuan, Chuanqi Tan, Wei Wang, Songfang Huang, and Fei Huang. 2023.
\newblock Rrhf: Rank responses to align language models with human feedback without tears.
\newblock \emph{arXiv preprint arXiv:2304.05302}.

\bibitem[{Zhang et~al.(2021)Zhang, Ju, Zhu, He, Shao, Zheng, and Liu}]{zhang2021co}
He~Zhang, Fusong Ju, Jianwei Zhu, Liang He, Bin Shao, Nanning Zheng, and Tie-Yan Liu. 2021.
\newblock Co-evolution transformer for protein contact prediction.
\newblock \emph{Advances in Neural Information Processing Systems}, 34:14252--14263.

\bibitem[{Zhang et~al.(2022)Zhang, Bi, Liang, Cheng, Hong, Deng, Lian, Zhang, and Chen}]{zhang2022ontoprotein}
Ningyu Zhang, Zhen Bi, Xiaozhuan Liang, Siyuan Cheng, Haosen Hong, Shumin Deng, Jiazhang Lian, Qiang Zhang, and Huajun Chen. 2022.
\newblock Ontoprotein: Protein pretraining with gene ontology embedding.
\newblock \emph{arXiv preprint arXiv:2201.11147}.

\bibitem[{Zhou et~al.(2024)Zhou, Zheng, Wu, Yi, Zhong, Tan, Liu, Li{\`o}, and Hong}]{zhou2024conditional}
Bingxin Zhou, Lirong Zheng, Banghao Wu, Kai Yi, Bozitao Zhong, Yang Tan, Qian Liu, Pietro Li{\`o}, and Liang Hong. 2024.
\newblock A conditional protein diffusion model generates artificial programmable endonuclease sequences with enhanced activity.
\newblock \emph{Cell Discovery}, 10(1):95.

\bibitem[{Zimmer(2002)}]{zimmer2002green}
Marc Zimmer. 2002.
\newblock Green fluorescent protein (gfp): applications, structure, and related photophysical behavior.
\newblock \emph{Chemical reviews}, 102(3):759--782.

\end{thebibliography}
\newpage

\appendix
\onecolumn

\section{More Details for Methods}\label{ap:dm}

\subsection{Construction of PSKG}
\label{ap:PSKG}
The PSKG leverages the structured biological knowledge encoded in the Gene Ontology, which provides a hierarchical vocabulary of GO terms describing biological processes, cellular components, and molecular functions. These GO terms $G = \{g_1,g_2,...,g_Z\}$ serve as nodes within the graph, connected by GO-GO triples, $(g_i,r,g_z)$, where $r$ represents relationships such as ``is-a'' or ``part-of''. 
These triples define hierarchical and associative relationships between terms, forming the backbone of the graph. The PSKG also incorporates associations between proteins and GO terms through Protein-GO triples, $(p,r,g)$, where $p \in P_\text{H} \cup P_\text{B}$ represents harmful or benign proteins and $g \in G$ is the associated GO term. These triples encode functional annotations that link proteins to their biological roles, establishing indirect relationships between $P_\text{H}$ and $P_\text{B}$. The harmful protein $P_\text{H}$ dataset was curated from UniProt by searching for sequences annotated with the keywords "toxin" (approximately 10,000 entries) and "antigen" (approximately 8,000 entries). After removing duplicates, the remaining sequences formed the harmful protein set. The benign protein dataset $P_\text{B}$ was collected from Swiss-Prot by excluding proteins identified as harmful.
By integrating these two layers—GO-GO relationships and Protein-GO associations—the PSKG models complex dependencies across the biological network. For instance, harmful proteins $P_\text{H}$ and benign proteins $P_\text{B}$ may share overlapping functional annotations, such as catalytic activity or molecular binding specificity, but harmful proteins often exhibit additional properties that contribute to harmful effects. 
\subsection{Algorithm definition}
\label{ap:Algorithm}
We provide the pseudo code of node pruning with weighted metrics as follows so that readers can easily understand the whole process.



    

    




    





\begin{algorithm}[!h]
\SetAlgoLined
\textbf{Input:} PSKG \ensuremath{G = (V, E)}, 
Set of harmful proteins \ensuremath{P_H}, benign proteins \ensuremath{P_B}, Gene Ontology (GO) nodes \ensuremath{G}, 
pruning thresholds \ensuremath{K, Q}, weight parameters \ensuremath{\alpha, \beta, \gamma, \delta}.

\textbf{Output:} Pruned graph \ensuremath{G' = (V', E')}

\ForEach{\ensuremath{g_z \in G}}{
    Compute harmful-benign bridging degree: 
    \ensuremath{
    R(g_z) = \sum_{p_H^i \in P_H} \sum_{p_B^j \in P_B} \mathbf{1}((p_H^i, g_z) \in E) \cdot \mathbf{1}((g_z, p_B^j) \in E)
    }

    Compute neighbor breadth: 
    \ensuremath{
    O(g_z) = \sum_{p_B^j \in P_B} \mathbf{1}((g_z, p_B^j) \in E)
    }

    Compute GO node importance: 
    \ensuremath{
    C(g_z) = \gamma \cdot R(g_z) + \delta \cdot O(g_z)
    }
}

Select the top-\ensuremath{Q} GO nodes with highest \ensuremath{C(g_z)} values:
\ensuremath{
G_{\hat{h}} \gets \mathrm{Top}\text{-}Q\ \mathrm{GO\ nodes\ based\ on\ } C(g_z)
}

\ForEach{\ensuremath{p_B^j \in P_B}}{
    Compute GO association factor:
    \ensuremath{
    C_{\mathrm{GO}}(p_B^j) = \sum_{g_z \in G_{\hat{h}}} \mathbf{1}((p_B^j, g_z) \in E)
    }

    Compute degree centrality:
    \ensuremath{
    C_{\mathrm{Deg}}(p_B^j) = \sum_{v \in V} \mathbf{1}((p_B^j, v) \in E)
    }

    Compute final weighted importance score:
    \ensuremath{
    S(p_B^j) = \alpha \cdot C_{\mathrm{GO}}(p_B^j) + \beta \cdot C_{\mathrm{Deg}}(p_B^j)
    }
}

Select top-\ensuremath{K} benign protein nodes with highest \ensuremath{S(p_B^j)} values:\\
\ensuremath{
P_B' \gets \mathrm{Top}\text{-}K\ \mathrm{benign\ protein\ nodes\ based\ on\ } S(p_B^j)
}

Define pruned graph: 
\ensuremath{
V' = P_H \cup P_B' \cup G_{\hat{h}}, \quad E' = \{(u, v) \in E \mid u, v \in V'\}
}

Return pruned graph \ensuremath{G' = (V', E')}.
\caption{Graph pruning algorithm based on PSKG}
\end{algorithm}

\subsection{Pruning threshold selection}\label{ap:prune}

\begin{figure*}[h]
    \centering
    \includegraphics[width=\columnwidth]{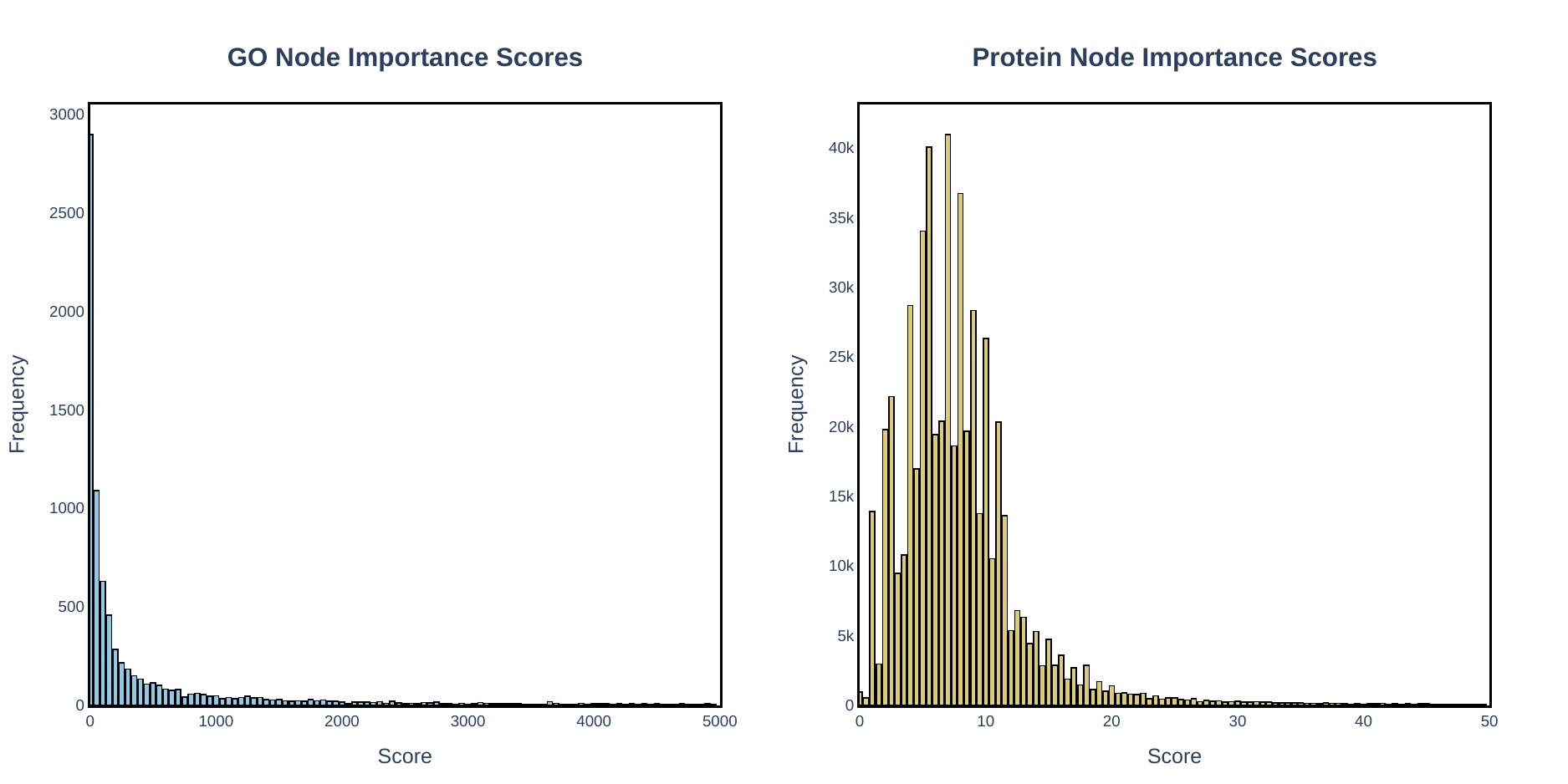}
    \vspace{-1em}
    \caption{Visualization of importance scores of GO nodes and protein nodes.}
    \label{fig:Importance_scores}
\end{figure*}

Figure \ref{fig:Importance_scores} presents the distribution of importance scores for both GO nodes and protein nodes within the PSKG. The importance scores were computed using the weighted metrics described in the methodology, which integrate structural and functional relevance factors to quantify the significance of each node. The figure clearly illustrates that the importance scores are highly unevenly distributed, with a significant proportion of nodes exhibiting low scores, while a smaller subset of nodes has substantially higher scores.

To optimize the pruning process, we selected the top-$Q=50\%$ of GO nodes and top-$K=50\%$ of protein nodes based on their importance scores. This threshold ensures a balance between retaining the most biologically significant nodes and reducing the size of the graph to improve computational efficiency. The decision to use the $50\%$ threshold is grounded in the observed score distributions. For GO nodes, the distribution shows a steep decline in the frequency of nodes with higher scores, indicating that the top-scoring nodes capture the majority of the functional and structural relevance in the graph. Similarly, for protein nodes, the score distribution exhibits a long-tailed pattern, with the top $50\%$ of nodes containing the majority of high-impact nodes based on the computed importance metrics. The choice of the $50\%$ threshold is further justified by the diminishing returns observed in the importance scores beyond this point. Nodes with scores below the median contribute marginally to the biological relationships within the graph, as their connections to key proteins or GO terms are sparse or weak. Retaining these nodes would unnecessarily inflate the graph size without providing meaningful contributions to downstream tasks.
The importance scores for GO nodes are calculated using weighting parameters $\gamma=1.0$  and $\delta=0.5$, while the protein node scores are determined using  $\alpha=0.5$ and $\beta=0.5$.These values were empirically selected to balance the contributions of key structural and relational factors in the graph. 

\section{Experimental setup}\label{ap:ex}

\paragraph{Baseline} We consider three foundational models and conducted a comprehensive comparative analysis. 
\begin{itemize}
\item \textbf{ProtGPT2} ProtGPT2 is a generative PLM based on the GPT-2 architecture, designed specifically for protein sequence generation. The model is pre-trained on a large corpus of protein sequences from databases like UniProt and SwissProt, using an autoregressive objective to predict the next token in a sequence. This approach allows the model to capture sequential dependencies and generate biologically plausible protein sequences with high fidelity. ProtGPT2 excels in creating diverse sequences with well-formed primary structures, making it a widely adopted baseline for protein generation tasks. 
\item \textbf{Progen2} Progen2 is a PLM pre-trained using a transformer-based architecture, leveraging a large-scale masked language modeling (MLM) objective. Unlike autoregressive models like ProtGPT2, Progen2 focuses on learning bidirectional contextual representations, making it particularly effective in capturing long-range dependencies within protein sequences. This capability allows Progen2 to perform well in tasks requiring an understanding of global sequence context, such as predicting functional domains or secondary structures.
\item \textbf{InstructProtein} InstructProtein is a specialized PLM designed for instruction-based generation tasks. It incorporates a pre-training strategy that combines supervised fine-tuning with reinforcement learning from human feedback (RLHF), enabling the model to follow specific generation instructions provided during inference. The instruction-based approach enhances the model's controllability and makes it particularly suited for applications requiring specific functional or structural characteristics in generated proteins.
\end{itemize}

\paragraph{Evaluation indicators}
\begin{itemize}
\item \textbf{Safety Evaluation Metrics}. In our evaluation, BLAST (Basic Local Alignment Search Tool) was employed to align the generated protein sequences with a curated testing harmful protein dataset for evaluation. Key metrics such as sequence identity, E-values, and alignment length were analyzed, as high sequence identity and low E-values often indicate strong functional or evolutionary relationships. Such relationships suggest a higher likelihood that the generated protein might share harmful properties with known harmful proteins. Complementing this, MMseqs2 was utilized for high-throughput sequence similarity searches and clustering. MMseqs2 provided additional insights by grouping similar sequences and uncovering potential functional associations between the generated proteins and harmful sequences.

To further assess potential harmfulness, we analyzed domain-level similarities using two approaches within the Pfam database. In the dynamic threshold method, a match was considered significant if the E-value of a domain in the generated protein was lower than the threshold E-value of the corresponding domain in the harmful protein database. This approach accounts for the unique statistical characteristics of each domain, as each hidden Markov model (HMM) in Pfam has its own noise model. By tailoring the significance criteria to individual domains, the dynamic threshold method provides a more precise and domain-specific evaluation of potential harmful associations.

In contrast, the fixed threshold method applied a standardized E-value cutoff across all domains, which we set to 0.001. This method simplifies the evaluation process by enforcing a consistent significance level, ensuring robust and highly significant matches are prioritized. To be considered a match under this method, both the E-value of the domain in the generated protein and the corresponding E-value in the harmful protein database had to fall below the fixed threshold. This ensures consistency while maintaining high statistical confidence in the results, allowing for a systematic evaluation of domain overlap across protein families.

\item \textbf{Functional Evaluation Metrics}. The GB1, PhoQ, UBC9, and GFP datasets represent diverse protein families, each with distinct functional roles in biological processes, providing a robust framework for evaluating the fine-tuned model’s performance. By assessing these datasets, we ensured that the fine-tuning process using KPO improved safety by reducing harmful protein generation without compromising the model’s ability to predict biologically relevant and functionally advantageous mutations. The evaluation relied on two key metrics: the generation probability score, which reflects how likely the model is to generate a particular mutation based on its learned patterns, and fitness, which measures how well a mutation preserves or enhances the protein's intended function. By comparing the average fitness of the top-ranked mutations generated by both the pre-trained and fine-tuned models, we assessed whether KPO fine-tuning shifted the model’s capacity to identify mutations that optimize protein functionality.

The use of the GB1, PhoQ, UBC9, and GFP datasets validated the dual objectives of the KPO method: improving safety while maintaining or enhancing functionality. This comprehensive evaluation framework demonstrated that KPO fine-tuning effectively aligns the model’s outputs with biological safety and functional integrity, ensuring the generation of proteins that are both safe and biologically meaningful.
\end{itemize}

\paragraph{Implementation Details} Our method was implemented using the PyTorch deep learning framework. To support reinforcement learning during the fine-tuning process, we integrated the TRL~\cite{vonwerra2022trl}. All experiments were conducted on an Ubuntu server with 8 NVIDIA A100 GPUs, each with 40GB of memory. The fine-tuning process was performed on protein language models with varying parameter scales: ProtGPT2 (738M parameters), ProGen2 (764M parameters), and InstructProtein (1.3B parameters). Training each epoch on our A100 GPUs required approximately 2 hours, with the learning rate set around 0.00005 to ensure stable and effective optimization. The dataset of harmful protein sequences, curated from UniProt, was split into training and testing sets in an 8:2 ratio. The training dataset was used to construct the PSKG, forming the harmful protein nodes that guide the fine-tuning process. The testing dataset served as the benchmark for evaluating sequence similarity and other safety-related metrics in the experimental results. Notably, the incorporation of the graph pruning algorithm significantly reduced the time required for generating preference data, with the overall time being halved compared to the baseline. This improvement in computational efficiency is attributed to the pruning process, which reduces the complexity of the knowledge graph by retaining only the most critical nodes and edges, thus accelerating downstream tasks such as preference data generation without compromising the quality of the generated outputs.

\begin{figure}[htbp] 
    \centering
    \includegraphics[width=1\textwidth]{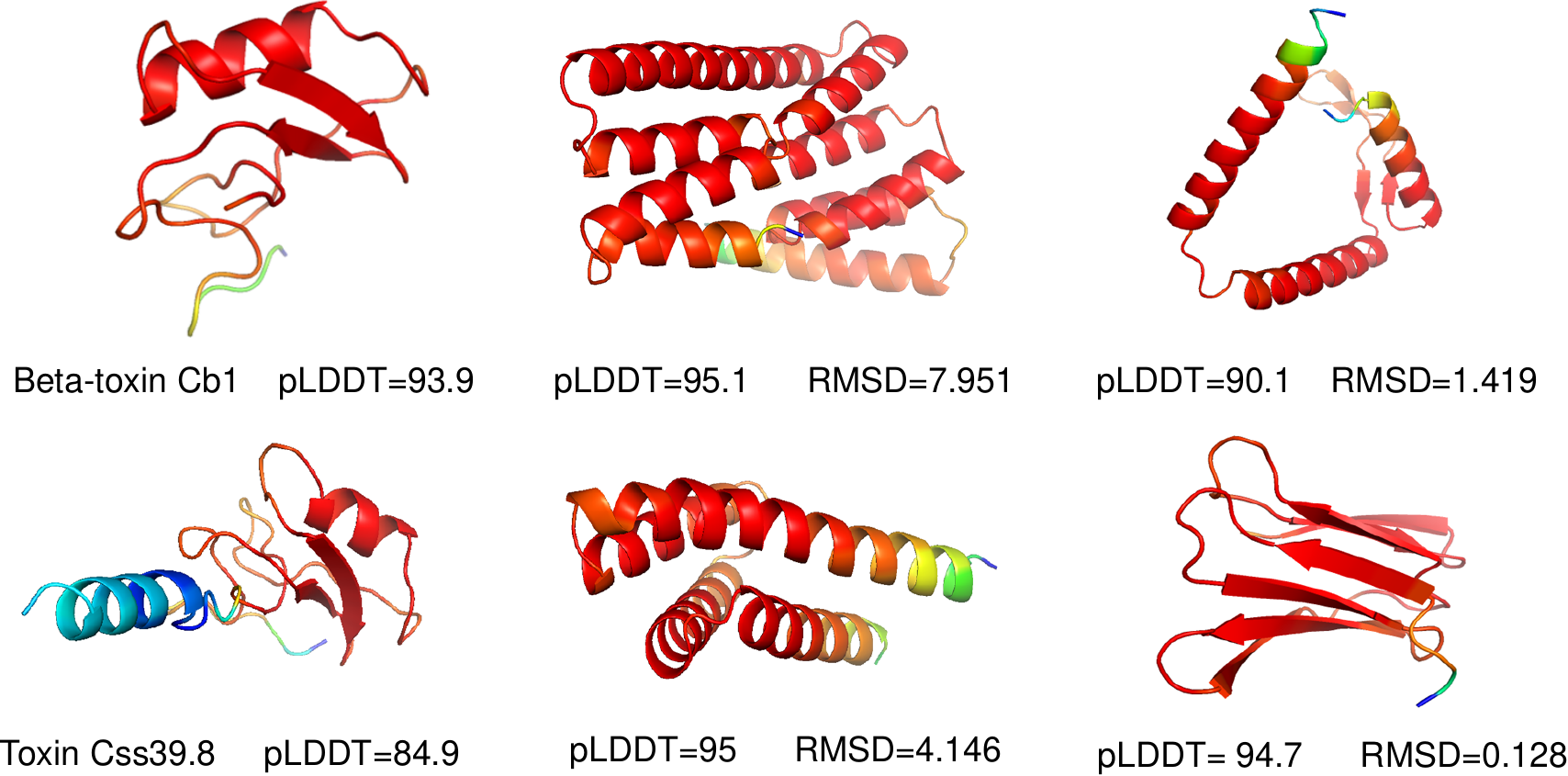}
    
    \vspace{0.5cm} 

    \includegraphics[width=1\textwidth]{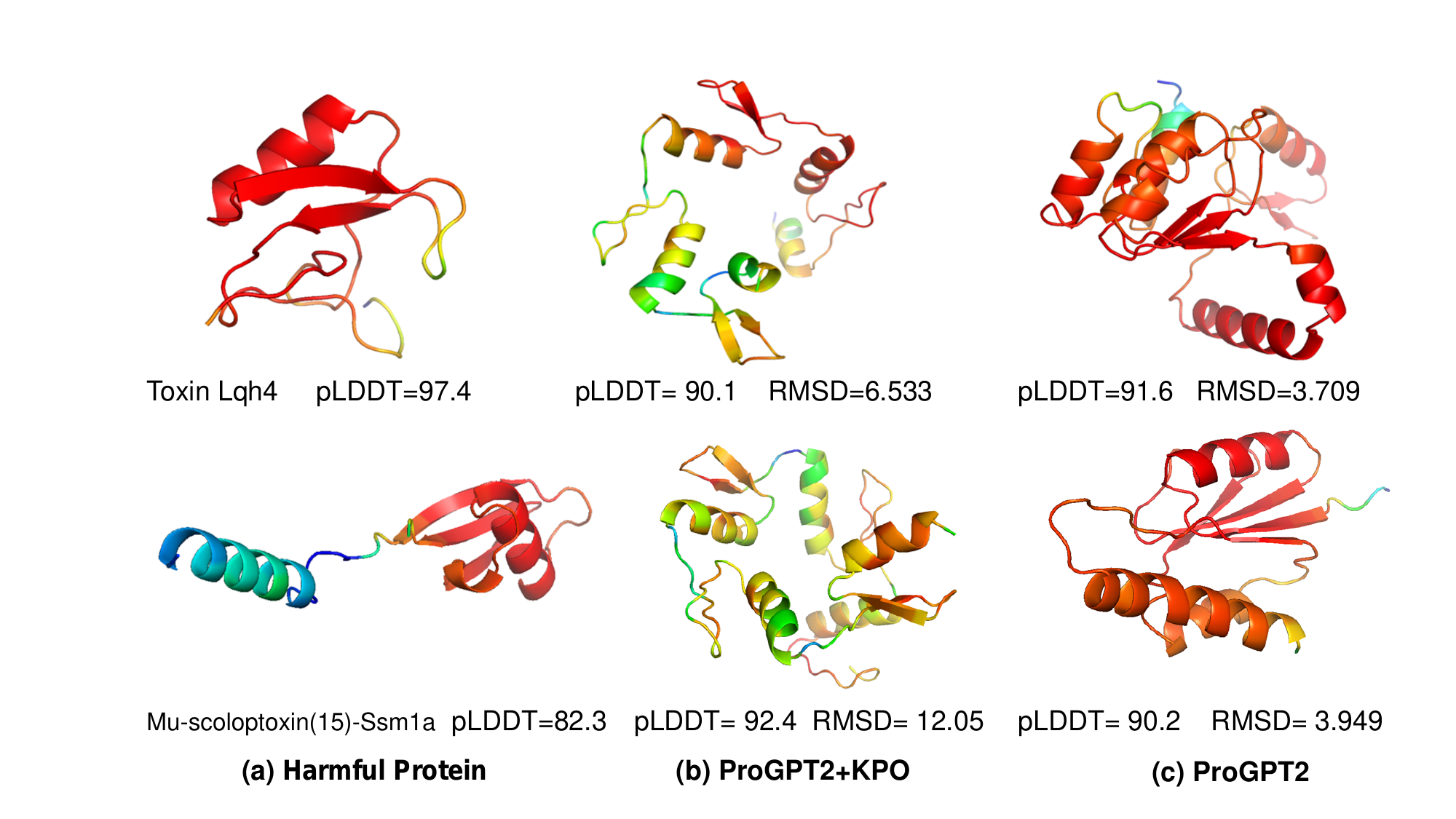}
    \caption{Comparison of 3D RMSD Values Between Harmful Proteins and Proteins Generated by Pre-trained and KPO-Fine-Tuned Models.}
    \label{fig:structure}
\end{figure}

\paragraph{\textbf{Three-dimensional Structure Analysis}} To evaluate the impact of the KPO fine-tuning method on the structural characteristics of generated proteins, we conducted a comparative analysis of the three-dimensional structures of proteins generated by the pre-trained and KPO-fine-tuned models. we employed ColabFold~\cite{mirdita2022colabfold,jumper2021highly} to predict the corresponding structural conformations of the generated proteins. For each selected protein, the predicted structure's confidence was evaluated using the mean predicted Local Distance Difference Test (pLDDT), which serves as a metric to quantify the reliability of the structural predictions.Meanwhile, we assessed the structural similarity between the generated proteins and known harmful proteins using Root Mean Square Deviation (RMSD).
Figure \ref{fig:structure} illustrates the comparison between harmful proteins and proteins generated by both the KPO-fine-tuned ProtGPT2 model and the pre-trained ProtGPT2 model. Each row of the figure corresponds to one of four harmful proteins with distinct functional and biochemical properties.The first harmful protein analyzed is a beta toxin that binds to sodium channels (Nav) at site-4 in a voltage-independent manner. By shifting the activation voltage to more negative potentials, this toxin enhances spontaneous and repetitive neuronal firing, which is lethal to mammals, including mice. The second beta toxin shares a similar mechanism of action but exhibits a broader specificity, affecting crustaceans and insects while being non-lethal to mammals. It also competitively displaces other beta toxins from mammalian sodium channels at higher concentrations. The third protein is an alpha toxin, binding to site-3 of sodium channels and inhibiting channel inactivation, leading to sustained neuronal activity. This toxin is highly harmful to both insects and mammals, causing convulsions and death upon injection into mice. Lastly, the fourth protein is a potassium channel blocker that targets the pore domain of Kv7 channel family members, disrupting ion transport and inducing severe physiological effects, including acute hypertension, coronary vasospasms, and seizures.
In all four cases, the KPO-fine-tuned ProtGPT2 model consistently generated proteins with significantly higher RMSD values compared to the pre-trained ProtGPT2 model. For example, when compared to the first beta toxin, the KPO-generated protein exhibited an RMSD of 7.951 Å, substantially higher than the 1.419 Å RMSD observed for the pre-trained model. Similar trends were observed across the remaining toxins, with RMSD values from the KPO-generated proteins exceeding those of the pre-trained model by approximately 50\% in most cases. These results indicate that the structural characteristics of the proteins generated by the KPO model diverge significantly from those of harmful proteins. By integrating safety knowledge from the PSKG, the KPO fine-tuning approach actively learns to generate proteins with sequence and structural features that diverge from those associated with harmfulness. This structural divergence is particularly critical as protein 3D structures are closely linked to their functional and interaction properties; reducing structural similarity to harmful proteins minimizes the risk of generating proteins with harmful biological activities.

In conclusion, the three-dimensional structure analysis provides strong evidence that the KPO fine-tuning approach significantly enhances the safety profile of protein generation models. By producing proteins with greater structural differences from harmful proteins, as evidenced by higher RMSD values, KPO fine-tuning ensures that the generated proteins are not only sequence-safe but also structurally distinct, advancing the biological safety and reliability of protein language models.

\paragraph{\textbf{Sensitivity Analyses of Hyperparameters}} \label{ap:Sensitivity}
The parameters $Q$ and $K$ determine the number of top-ranked GO nodes and benign protein nodes selected based on their importance scores. As $Q$ and $K$ decrease, pruning becomes more aggressive, which improves computational efficiency but may compromise safety in protein generation. As shown in Table~\ref{tab:qk_performance}, setting $Q = 50\%$ and $K = 50\%$ achieves the best trade-off between computational efficiency and safety, as reflected by improved performance on both safety and functional metrics.

\newcommand{\myfo}{\fontsize{9.8pt}{\baselineskip}\selectfont}

\begin{table}[!htb]\myfo
\centering
\caption{Performance under different $Q, K$ configurations.}
\begin{tabularx}{\textwidth}{@{\extracolsep{\fill}}p{1.5cm}cccccccccc@{\extracolsep{\fill}}}
\toprule
$\gamma$:$\sigma$ & BLAST$\mathord{\downarrow}$ & MMseq2$\mathord{\downarrow}$ & Pfam\_D$\mathord{\downarrow}$ & Pfam\_E$\mathord{\downarrow}$ & ToxinPred3$\mathord{\downarrow}$ & GB1$\mathord{\uparrow}$ & PhoQ$\mathord{\uparrow}$ & UBC9$\mathord{\uparrow}$ & GFP$\mathord{\uparrow}$  \\
\midrule
50\%, 50\% & 0.138 & 0.149 & \textbf{0.261} & \textbf{0.182} & \textbf{0.024} & \textbf{0.041} & 0.315 & 0.129 & \textbf{2.20} \\
30\%, 30\% & \textbf{0.115} & \textbf{0.103} & 0.338 & 0.397 & 0.045 & 0.037 & 0.329 & 0.142 & 1.72 \\
10\%, 10\% & 0.116 & 0.106 & 0.327 & 0.366 & 0.038 & 0.037 & \textbf{0.329} & \textbf{0.142} & 1.72 \\
\bottomrule
\end{tabularx}
\label{tab:qk_performance}
\end{table}

\section{Potential Risks}

The curated dataset of harmful protein sequences, although essential for constructing the PSKG and fine-tuning the model, could potentially be misused if accessed by malicious actors. Such misuse might involve training models specifically designed to generate even more harmful or weaponizable proteins, amplifying the risks to public health and ecological stability.To address these concerns, we have decided to either withhold public access to the harmful protein dataset or release it under strict guidelines and controlled conditions. By limiting its availability, we aim to balance the scientific value of this research with the imperative to minimize potential misuse, ensuring that the advancements introduced by KPO are applied responsibly in biotechnology and synthetic biology.

\end{document}